\documentclass[fullpaper]{nldl}

\renewcommand{\DOI}{12345}

\usepackage[utf8]{inputenc}
\usepackage{url}
\usepackage{graphicx}
\usepackage{authblk}

\usepackage{algorithm}
\usepackage{algorithmic}

\usepackage{natbib}[square,numbers]

\usepackage{hyperref}

\title{Point Cloud Data Simulation and Modelling\\ with Aize Workspace}
\author[]{Boris Mocialov, Eirik Eythorsson, Reza Parseh, Hoang Tran, Vegard Flovik}
\affil[]{Aize AS}

\date{\vspace{-5ex}}

\begin{document}
\nldlmaketitle

\begin{abstract}
This work takes a look at data models often used in digital twins and presents preliminary results specifically from surface reconstruction and semantic segmentation models trained using simulated data. This work is expected to serve as a ground work for future endeavours in data contextualisation inside a digital twin.
\end{abstract}

\vspace{-1em}
\section{Introduction}
Digital twin (DT) solutions promise real-time adaptation to internal or external stimuli and optimisation of operational processes by linking physical and digital entities together with the help of sensors and actuators. The sole purpose of a DT is to structure perceived information and to present it to its users who, in turn, might use this information for decision-making ~\cite{cimino2019review, roy2020digital, batty2018digital, van2020taxonomy}. 

The more complex the physical system, the more features its DT implements to fully mirror processes of that physical system. It is the case that often we see complex or very complex physical systems with many processes embedded in them. Therefore, deployed DT in industries often contain large collection of domain-specific expert knowledge, which needs to be applied to the information available to the DT. Although DT attempts to mirror complex industrial processes, the information that passes through DT needs to be presented to its user in a condensed non-overwhelming fashion.

This extended abstract focuses on semantic models necessary for DT to segment all the information that passes through it. These semantic models should not only classify which part of the physical system the data belongs to, but also what is the meaning of that data. A brief overview of various digital twin data models sets the mood before a use case supported by preliminary results followed by a discussion of digital twin models 'in the wild'.

\section{Digital Twin Data Models}
Data models are based on domain-specific taxonomic definitions and are designed to assist in turning information into structured data ~\cite{wang2010negotiation}. The two prominent methods for encoding knowledge digitally has been explicit rule specification lead by domain experts or implicit rule inference from data acquired in domain-specific context \cite{weiss2003knowledge}. Combination of the two methods also exists specifically in cases where availability of domain experts is limited \cite{xu2012feature, ratner2018snorkel, wang2020joining}. DT solutions tend to be in the crosshairs of the two contrary methods despite machine learning and deep learning attracting more interest \cite{tao2022digital}. 

Since IoT is one of the main drivers for development of DT solutions, these solutions are expected to support a wide range of data formats ~\cite{qi2021enabling}. Therefore, data models vary greatly depending on the data at hand employing such models as regression ~\cite{samnejad2020digital}, clustering ~\cite{priyanka2022digital}, deep learning ~\cite{razzaq2022deepclassrooms}, Markov chains ~\cite{ghosh2019hidden}, Monte Carlo ~\cite{chen2021digital}, and Gaussian processes ~\cite{chakraborty2021machine}. Data annotation is a very expensive stage in case of data-driven data models and the cost becomes even more significant as expert knowledge is required. Having said that data annotation for data-driven methods is costly, DT can provide simulation by design ~\cite{boschert2016digital}. User should be able to not only test hypotheses using DT through simulation ~\cite{kalidindi2022digital}, but also generate data for training data-driven models ~\cite{lee2022generating, mukhopadhyay2021generating} which has been a common approach in robotics for years ~\cite{pelossof2004svm, desai2020stochastic, zhu2020robosuite}. Although a powerful approach to data generation, simulation-reality gap is a known problem that has not been solved to this day ~\cite{jakobi1995noise}. Nevertheless, simulation is often used in transfer learning as a pre-training step ~\cite{weiss2016survey}, where only a limited amount of expert knowledge is required to fine-tune a model to a real-world domain. 

\section{Preliminary Results}
One of the main data sources in our DT is independently collected point cloud data from offshore platforms with an average point cloud density of approximately $40,000$ points per $m^3$. Available to us data currently does not have proper annotations, so we are generating synthetic point cloud data (currently without return strength of a laser beam) using digital twin of offshore platforms available as $3D$ models produced by CAD tools employed in the asset design phase. These $3D$ models available in the DT provide us free annotations for synthesised point cloud data. Obligatory data augmentation is used to generate noise in the data during training of machine learning and deep learning methods that i) semantically segment and ii) reconstruct surfaces.

\vspace{2em}
\begin{minipage}[t]{0.45\linewidth}
    \centering
    a)
\end{minipage}
\hfill
\begin{minipage}[t]{0.45\linewidth}
    \centering
    b)
\end{minipage}
\begin{figure}[htbp]
    \centering
    i) \includegraphics[width=0.45\linewidth]{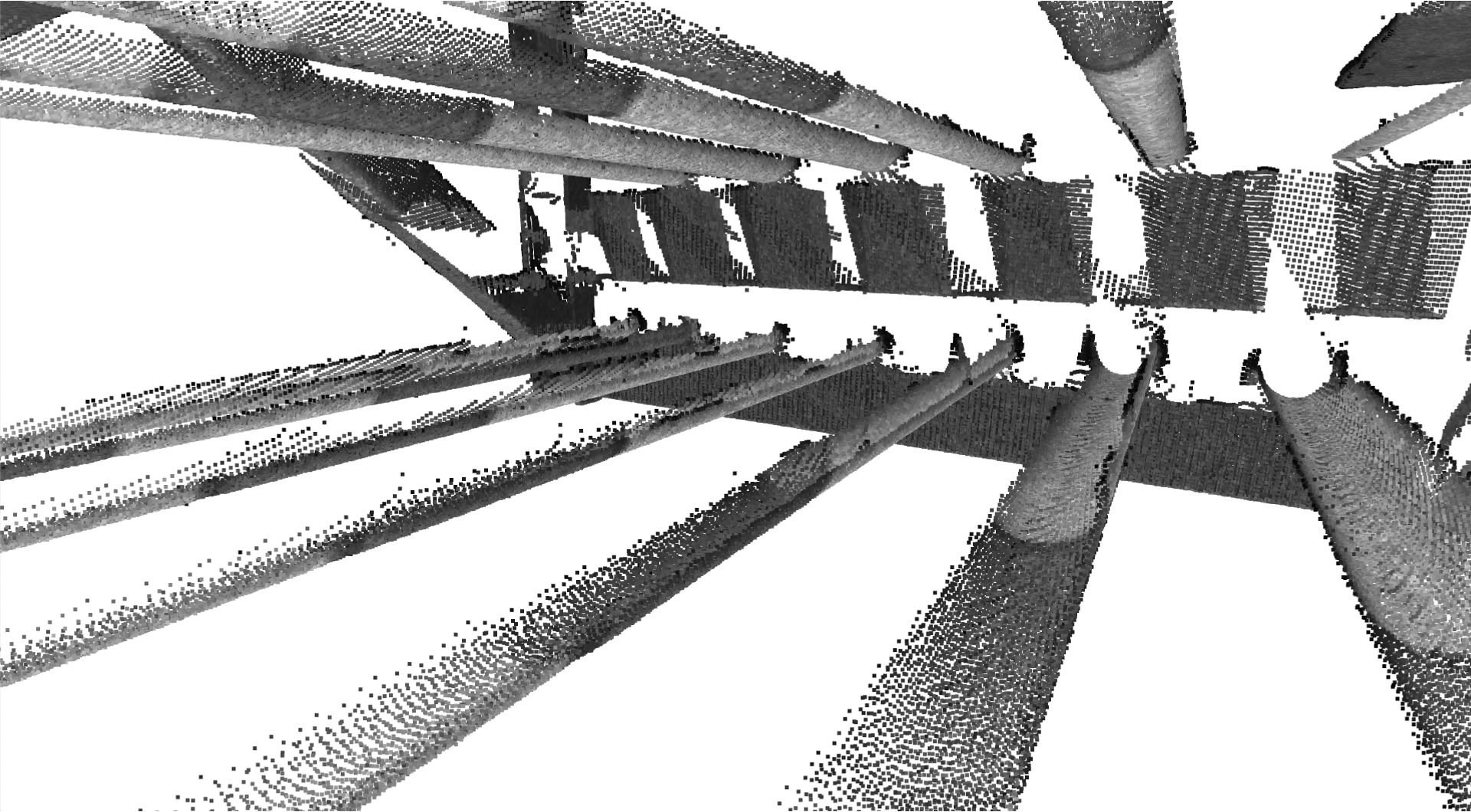}
    \includegraphics[width=0.45\linewidth]{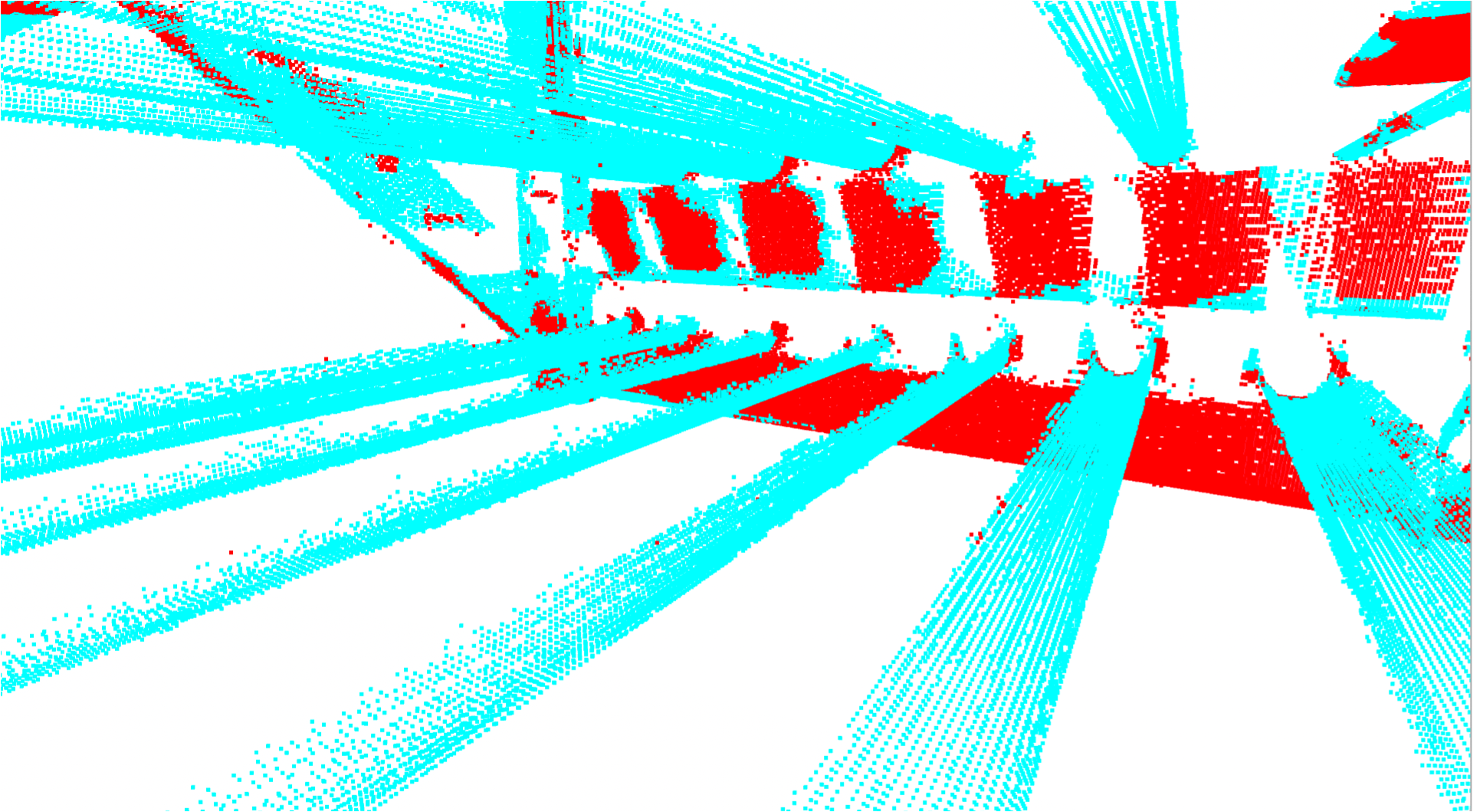}
    ii) \includegraphics[width=0.45\linewidth]{Picture1.png}
    \includegraphics[width=0.45\linewidth]{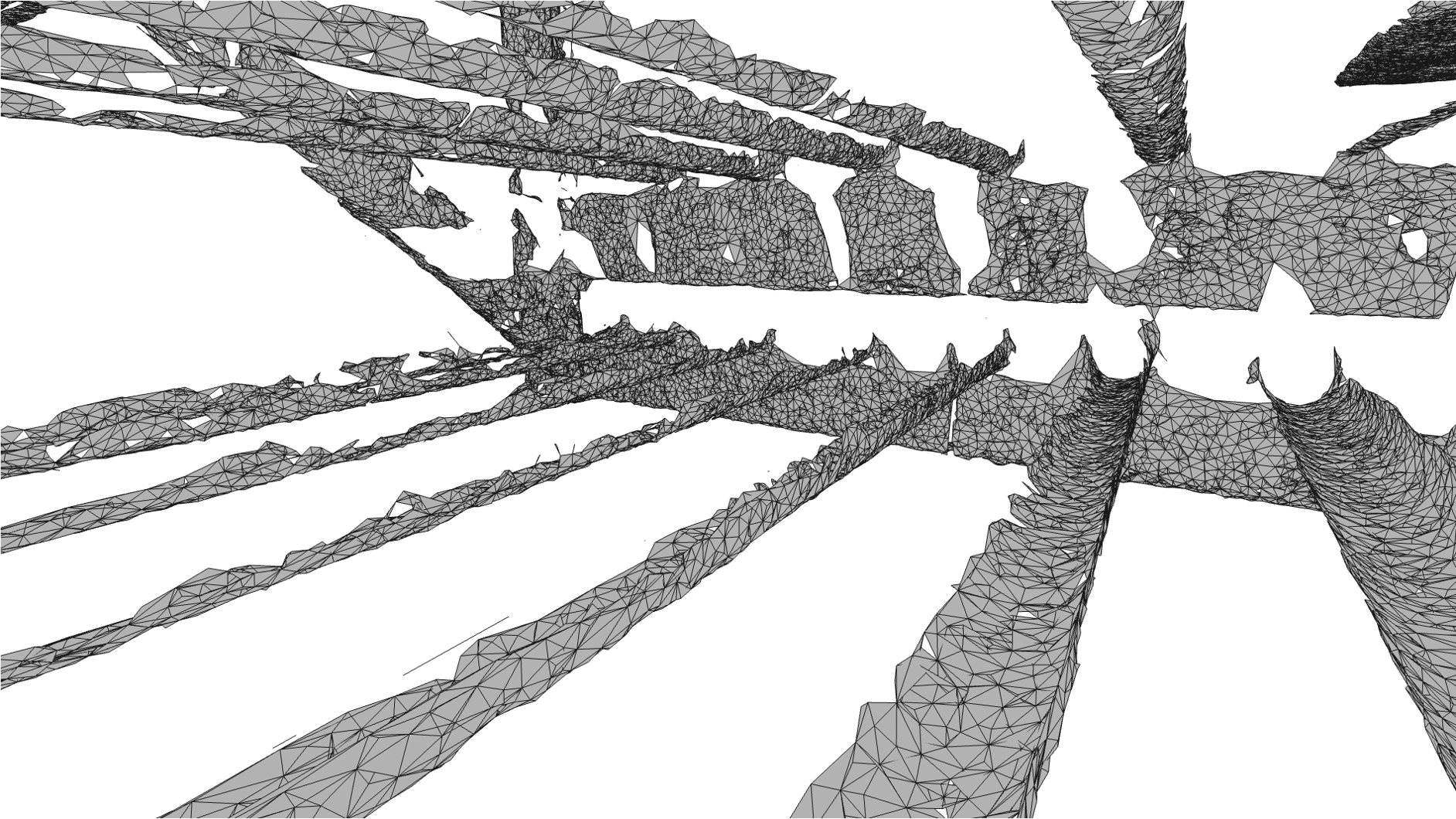}
    \caption{i) Semantic segmentation and ii) surface reconstruction models trained on simulated data using DT and tested on real offshore platform data. Column A) shows the same input portion of the real-world point cloud of an offshore platform and column B) shows the result after applying trained model to the raw data}
    \label{fig:pic1}
\end{figure}

Figure~\ref{fig:pic1} demonstrates preliminary results of two models trained on simulated point cloud data using offshore platform $3D$ models. Each row represent a different use case and a different model while column b) shows the result after a model has been applied to the raw data shown in column a). Row i) shows semantic segmentation use case, where each point in the raw point cloud is classified as either a pipe or not a pipe and row ii) shows surface reconstruction on the same raw data from column a).


\section{Discussion}
While results shown in Figure~\ref{fig:pic1} are promising and can be improved by adopting well-known tricks used in literature, the discussion should be aimed at the user of an expert tool such as a DT and tools interface. While it is possible to reduce false positives and improve model performance, we should not forget that the main job of a DT solution is to structure perceived information and present that information to its users. It is ultimately the user who should decide the meaning and correctness of predictions from all models employed by DT. Ultimately it comes to the human expert knowledge to judge the predictions and plan next steps. Models in DT, on the other hand, should be properly documented and maintained alongside the data that was used during training and fine-tuning. 

Moreover, models used by DT should be explainable especially when DT are used in safety-critical domains such as healthcare, transportation, defence, space, and oil\&gas. Such explinability could be achieved by performing ablation studies, which could uncover insights into behaviour of a model. Specifically, by reducing or removing parts of a complex model, it should be possible to identify responsibilities of different parts of complex models.

Spending more time on interpretation of trained models will provide additional confidence both from the developer and the user as well as higher level of control over trained models.

\section{Conclusion}
This extended abstract focused on semantic models necessary for DT to segment all the information that passes through DT and showed preliminary results obtained from surface reconstruction and semantic segmentation models trained on simulated data. Future work will explore further how data modelling can further be used in data contextualisation in DT.

\bibliographystyle{abbrvnat}
\bibliography{main}

\end{document}